\documentclass[10pt,twocolumn,letterpaper]{article}
\pdfoutput=1
\usepackage{iccv}
\usepackage{times}
\usepackage{epsfig}
\usepackage{graphicx}
\usepackage{arydshln}
\usepackage{amsmath}
\usepackage{amssymb}
\usepackage{subcaption}
\usepackage{booktabs}
\usepackage{multirow}
\usepackage{diagbox}
\usepackage{color}

\usepackage[title]{appendix}
\usepackage[colorlinks,linkcolor=red]{hyperref}

\iccvfinalcopy 

\def\iccvPaperID{****} 

\ificcvfinal\fi

\begin{document}

\title{Virtual Multi-Modality Self-Supervised Foreground Matting for \\
Human-Object Interaction}

\author{Bo Xu\textsuperscript{1}, Han Huang\textsuperscript{1}, Cheng Lu\textsuperscript{2},  Ziwen Li\textsuperscript{1,3} and Yandong Guo\textsuperscript{1,*}\\
\textsuperscript{1}OPPO Research Institute, \textsuperscript{2}Xmotors, \textsuperscript{3}University of California, San Diego\\
{\tt\small yandong.guo@live.com}
}

\maketitle
\ificcvfinal\thispagestyle{empty}\fi

\begin{abstract}
Most existing human matting algorithms tried to separate pure human-only foreground from the background. In this paper, we propose a Virtual Multi-modality Foreground Matting (VMFM) method to learn human-object interactive foreground (human and objects interacted with him or her) from a raw RGB image. The VMFM method requires no additional inputs, \eg trimap or known background. We reformulate foreground matting as a self-supervised multi-modality problem: factor each input image into estimated depth map, segmentation mask, and interaction heatmap using three auto-encoders. In order to fully utilize the characteristics of each modality, we first train a dual encoder-to-decoder network to estimate the same alpha matte. Then we introduce a self-supervised method: Complementary Learning(CL) to predict deviation probability map and exchange reliable gradients across modalities without label. We conducted extensive experiments to analyze the effectiveness of each modality and the significance of different components in complementary learning. We demonstrate that our model outperforms the state-of-the-art methods. Our code and data is available at \url{https://github.com/JackSyu/HOI-Matting} \vspace{-10pt}  
\end{abstract}

\section{Introduction}

Matting is known as a chroma keying process to separate the foreground from a single image or video stream and then composite it with a new background. It has long been applied in photography and special-effect film-making~\cite{xu2017deep}. Due to the rapid development of deep neural network in computer vision, automatic matting becomes increasingly matured~\cite{chen2013knn,levin2007closed,lu2019indices,sengupta2020background,xu2017deep,zhang2019late}. However, most existing deep learning matting methods extract pure human foreground under controlled settings.

\begin{figure}[t]
\centering
    \includegraphics[width=1.0\linewidth]{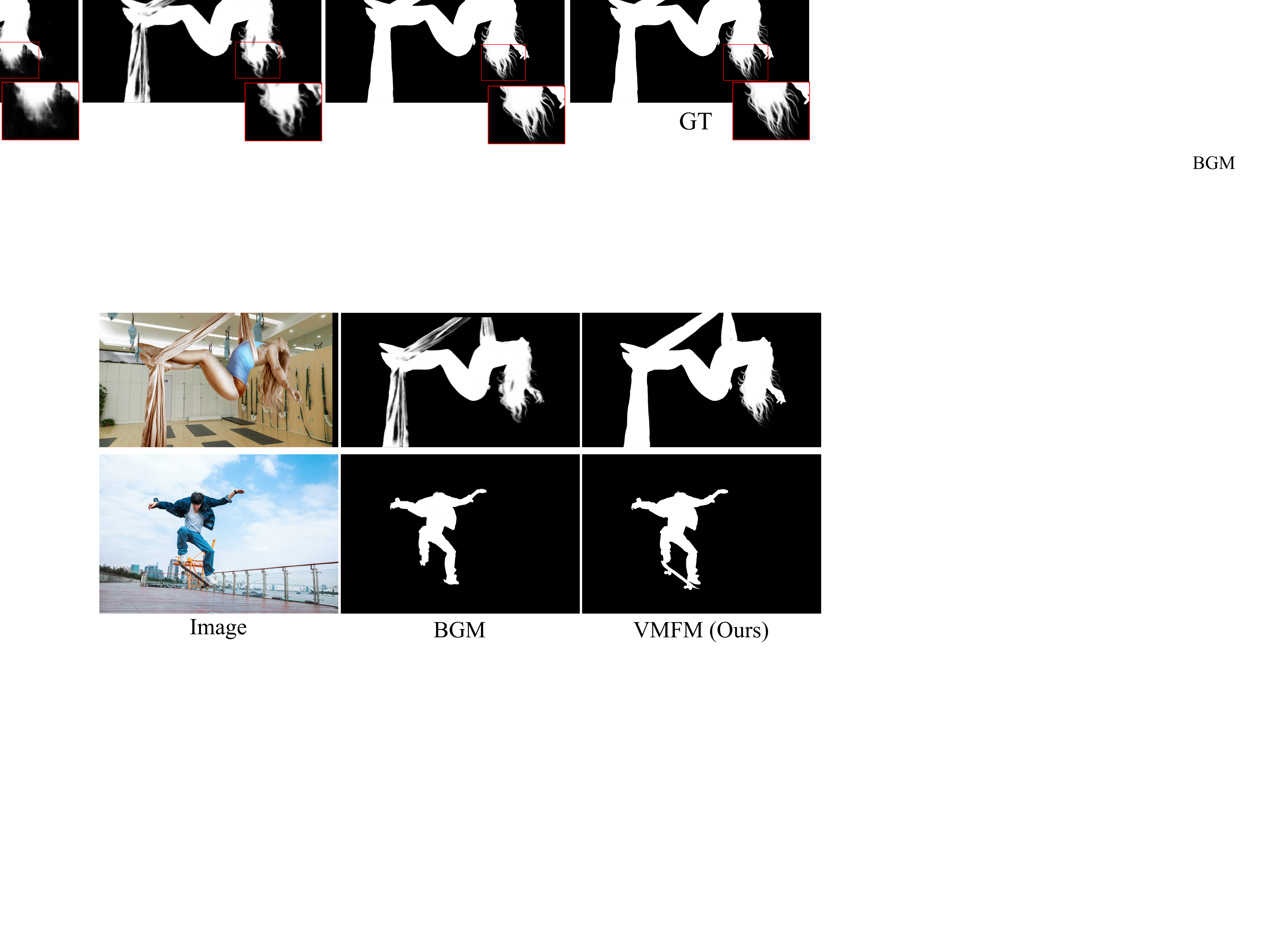}

\caption{Given challenging human-interactive images, one
of the recent state-of-the-art foreground matting approach
Background Matting (BGM)~\cite{sengupta2020background}, fails to produce accurate alpha mattes. However our proposed multi-modality model VMFM is able to separate accurate human-object interactive foreground mattes, outperforming SOTA works.} 
\label{fig:overall}
\vspace{-10pt}
\end{figure}

We revisit the matting problem with the following three beliefs. First, we argue that foreground matting algorithm should be able to handle situations when humans are interacting with objects. Currently, most of human matting algorithms~\cite{chen2018semantic,chen2013knn,levin2007closed,liu2020boosting,lu2019indices,sengupta2020background,zhang2019late} only focus on the body region while ignoring the object interacting with the person of interest, as illustrated in Figure~\ref{fig:overall}. Second,
the ideal matting model should be trained on a reliable and representative dataset, which requires the minimum amount of supervision and labeling effort. That is because the definition of good human-object interactive matting is very subjective and such labeling is typically expensive and difficult, if not impossible. Third, we believe that this human-object interactive foreground matting should be done under unconstrained conditions. Until now, most accurate matting techniques still rely on “blue-screen” or “green-screen” to remove background in a recording for traveling mattes. Some other methods require trimap (trimap is a draft marking foreground, background, and unknown areas) as an even strong prior~\cite{hou2019context,lu2019indices,xu2017deep,zhang2019late}. Learning without predefined background~\cite{sengupta2020background} or prior knowledge is critical.

To address the above issues, we propose the solution with following two components accordingly. First, we reformulate this problem as a multi-modality task to better capture both human of interest and the interacted objects. We believe feature extracted in the RGB space is not enough to produce high quality mattes. In our case, we introduce depth estimation as the extra virtual modality on top of RGB image and estimated segmentation mask. We argue that while RGB and segmentation map is good at predicting human contours and regions with similar color (skin \etal)~\cite{sengupta2020background}, however it struggles to recognize the association between human and the interacted object. Compared with a human body, the interacted objects are typically more diverse in terms of appearance \eg on shape, color, and type. That makes detection of the human-object association based on similar appearance in RGB space almost impossible. However, the depth modality can easily group human and object into one entity because of the high degree of continuity in its feature space. However, most current segmentation networks are not good at finding depth-related feature since they are trained intentionally to separate, rather than to group, different objects. To avoid ambiguity, we follow the definition of interaction as in~\cite{gupta2015visual}, which specifies human-object interaction into 26 classes. To make our model focus on the features of humans and interactive objects, we introduce interaction heatmaps which are also generated from RGB image as the semantic prior.   

\begin{figure*}[t]
\centering
    \includegraphics[width=0.92\linewidth]{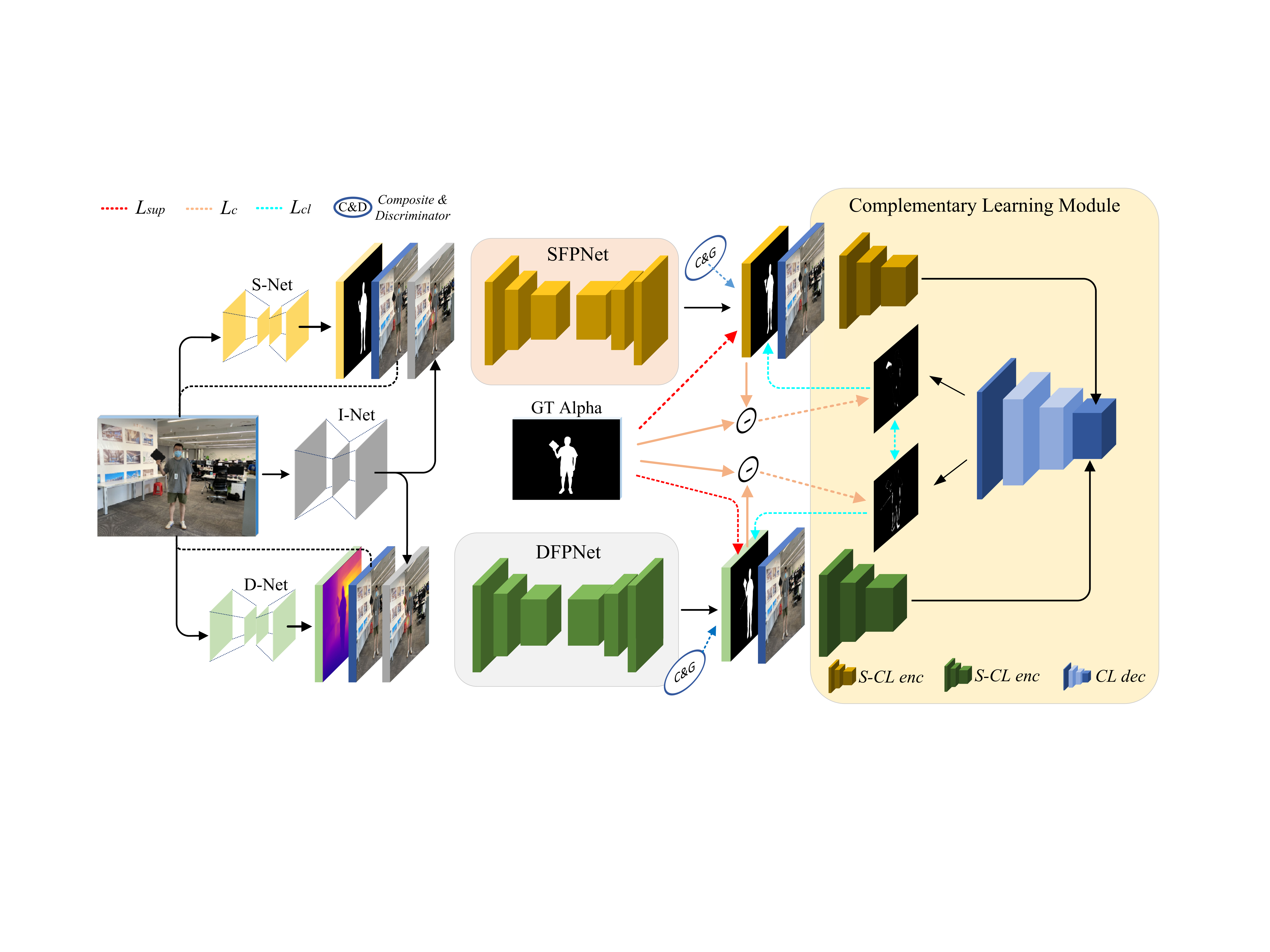}

\caption{Architecture of the Virtual Multi-modality Foreground Matting ({\bf VMFM}) network in the training stage. The Foreground Prediction ({\bf FP}) module consists of dual networks {\bf SFPnet} and {\bf DFPnet}, which receives different modalities as inputs and estimate the same alpha matte. The Complementary Learning ({\bf CL}) module consists of two encoders {\bf S-CL enc}, {\bf D-CL enc} and one decoder {\bf CL dec}, outputting deviation probability maps of the predicted alpha mattes.}

\label{fig:architecture}
\end{figure*}

Secondly, to minimize the labeling cost we propose the Complementary Learning (CL) self-supervised module in our VMFM network. The output of each modality can supervise or be supervised by its counterpart. Collectively, they manage to find the best combination. That is inspired by the fact each modality has its unique weakness and strength. For example in Figure~\ref{fig:architecture}, depth modality can help us predict the association between a man and the book in his hand but is weak on distinguishing the outline between ground and feet. On the contrary, color feature can easily tell the feet from the ground but struggle to know that the book is part of the foreground. However, we can learn a collection of more reliable elements from both modalities through pixel-wise complementary learning; see Figure~\ref{fig:results} for examples of VMFM's outputs.

The VMFM network consists of two stages: Foreground Prediction (FP) and Complementary Learning (CL). The FP stage includes two modules: segmentation-based foreground prediction network (SFPNet) and depth-based foreground prediction network (DFPNet), both in encoder-to-decoder fashion. The depth map is estimated by a depth estimation network (D-Net) from a single RGB image and the network is based on the backbone of~\cite{alhashim2018high}. The I-Net, in which Hourglass~\cite{newell2016stacked} backbone embeds the human-object interactive feature, is used to guide the matting by generating attentional heatmaps. 
In the CL stage, each output pixel of one FP network can supervise, or be supervised by its counterpart (output pixel at the same location in the other network), depending on which one has a lower deviation probability. Jointly, we minimize the unreliable pixels (probability higher than the threshold we set) to better highlight the foreground pixels that belong to either human or interacted objects.                
To justify our solutions, we compare our algorithm, objectively and subjectively with other methods. Also, we demonstrate by ablation study that different modalities focus on different, if not unique, details. Complementary learning can combine their reliable pixels while suppressing unreliable ones. Overall, the contributions of this paper are as follows:
\begin{itemize}
    \item This is the first end-to-end foreground/alpha matting algorithm that focuses on human-object interactive scenarios under unconstrained conditions.
    \vspace{-6pt}
    \item We reformulate foreground alpha matting as a cross-modal self-supervised task, which minimizes the labeling cost and produces fine matting by leveraging both modalities.
    \vspace{-6pt}
    \item Extensive experiments demonstrate the effectiveness of VMFM, outperforming the state-of-the-art (SOTA) methods in real-world scenarios.
    \vspace{-6pt}
\end{itemize}    

\section{Related works}

Currently, matting is generally formulated as an image composite problem, which solves the 7
unknown variables per pixel from only 3 known values:
\begin{equation}
    I_{i} = \alpha_{i} F_{i} + (1-\alpha_{i})B_{i}
\label{E0}
\end{equation}
where 3 dimensional RGB color $I_{i}$ of pixel $i$, while foreground
RGB color $F_{i}$, background RGB color $B_{i}$ and matte estimation $\alpha_{i}$ are unknown. In this section, we discuss the SOTA works dealing with this under-determined equation. 

\subsection{Classic methods}
Classic foreground matting methods can be generally categorized into two approaches: sampling-based and propagation-based. Sampling-based methods~\cite{aksoy2017designing,chen2013image,shahrian2013improving,wang2007optimized,he2011global,chuang2001bayesian} sample the known foreground and background color information, and then leverage them to achieve matting in the unknown region. Various sampling-based algorithms are proposed, \eg Bayesian matting~\cite{chuang2001bayesian}, optimized color sampling~\cite{wang2007optimized}, global sampling method~\cite{he2011global} and comprehensive sampling~\cite{shahrian2013improving}. Propagation-based methods~\cite{levin2008spectral,chen2013knn,he2010fast,lee2011nonlocal,levin2007closed,sun2004poisson} reformulate the composite Eq.~\ref{E0} to propagate the alpha values from the known foreground and background into the unknown region, achieving more reliable matting results.~\cite{wang2008image} provides a very comprehensive review of different matting algorithms. 

\subsection{Deep learning based methods}
Classic matting methods are carefully designed to solve the composite equation and its variant versions. However, these methods heavily rely on chromatic information, which leads to bad quality when the color of foreground and background shows small or no noticeable difference.

Automatic and intelligent matting algorithms are emerging, due to the rapid development of deep neural network in computer vision. Initially, some attempts were made to combine deep learning networks with classic matting techniques, \eg closed-form matting~\cite{levin2007closed} and KNN matting~\cite{chen2013knn}. Cho \etal~\cite{cho2016natural} employ a deep neural network to improve results of the closed-form matting and KNN matting. These attempts are not end-to-end, so not surprisingly the matting performance is limited by the convolution back-ends. 
Subsequently, full DL image matting algorithms appear~\cite{chen2018semantic,chen2018tom,xu2017deep}. Xu \etal~\cite{xu2017deep} propose a two-stage deep neural network (Deep Image Matting) based on SegNet~\cite{badrinarayanan2017segnet} for alpha matte estimation and create a large-scale image matting dataset (Adobe dataset) with ground truth foreground (alpha) matte, which can be composited over a variety of backgrounds to produce training data. And we also use this data for the first-step pre-training of our network. Lutz \etal ~\cite{lutz2018alphagan} introduce a generative adversarial network (GAN) for natural image matting and improve the results of Deep Image Matting~\cite{xu2017deep}. Cai \etal ~\cite{cai2019disentangled} investigated the bottleneck of the previous methods that directly estimate the alpha matte from a coarse trimap, and propose to divide matting problem into trimap adaptation and alpha estimation tasks. Hou \etal ~\cite{hou2019context} employs two encoder networks to extract essential information for matting, however it is not robust to faulty trimaps.             

{\bf Trimap-free methods.} Currently, a majority of deep image matting algorithms~\cite{cai2019disentangled,hou2019context,lutz2018alphagan,xu2017deep} try to estimate a boundary that divides the foreground and background, with the aid of a user-generated trimap. Several trimap-free matting methods~\cite{chen2018semantic,zhang2019late} predict the trimap first, followed by alpha matting. 

{\bf Additional natural background.} Qian \etal~\cite{qian1999video} compute a probability map to classify each pixel into the foreground or background by simple background
subtraction. This algorithm is sensitive to the threshold and fails when the colors of foreground and background are close. Sengupta \etal~\cite{sengupta2020background} introduce a self-supervised adversarial approach - Background Matting (BGM), achieving state-of-the-art results. However, a photographer needs to take a shoot of natural background first, which is not friendly to the intensive multi-scene shooting application.

Besides, current DL foreground matting methods focus only on human matting and ignore the fact that objects interacting or attached with people can always, if not every time, be part of the foreground. This is the main reason why we propose the method of VMFM. In this paper, we quantitatively evaluate the performance of our model for alpha matting in human-object interactive scenes. 

\begin{figure}[t]
\centering
    \includegraphics[width=1.0\linewidth]{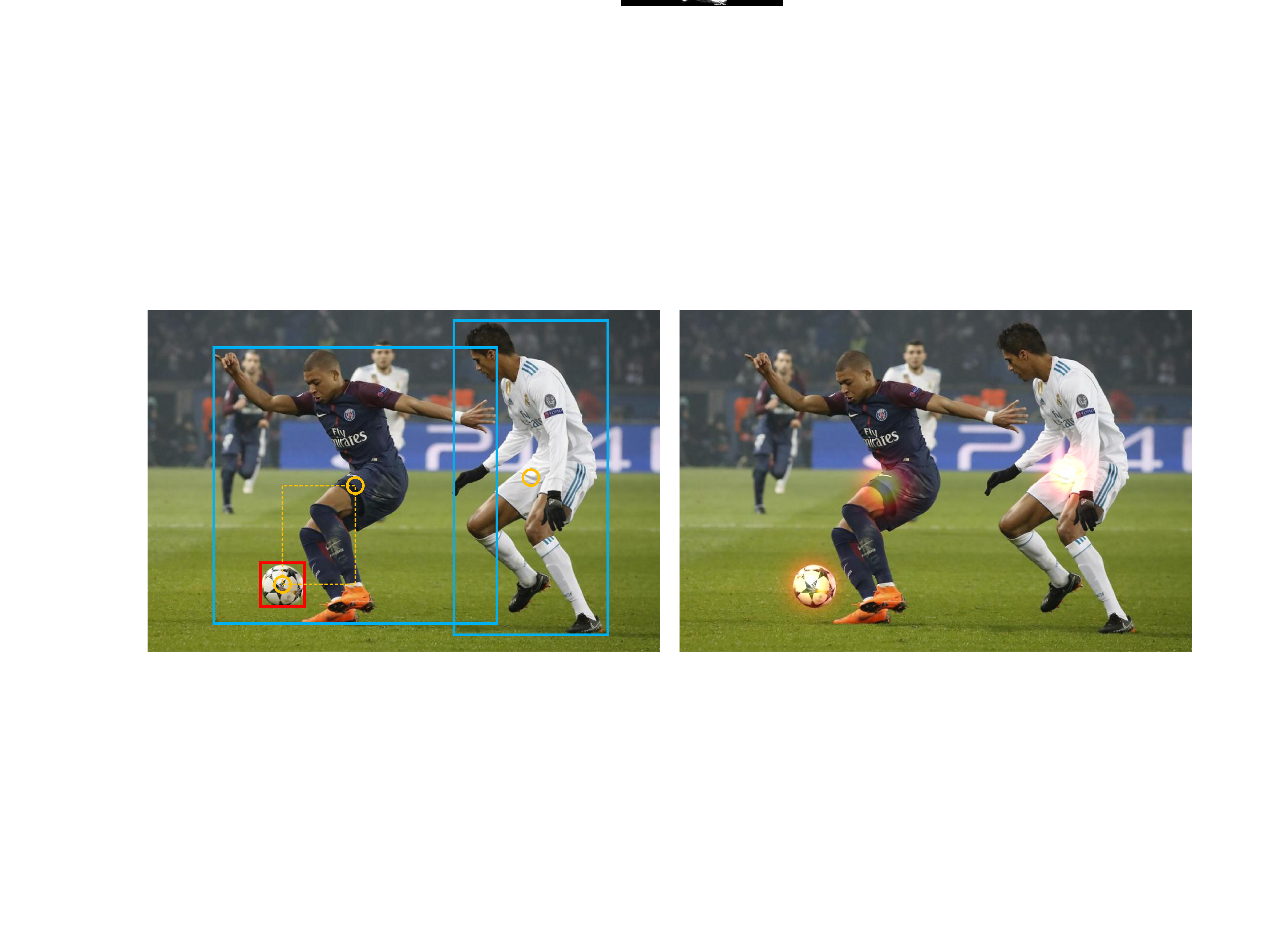}

\caption{Illusion of interaction pair heatmaps on an example image. Interaction pair means human and one of the objects interacted with him or her. Classes of such interactions are predefined and cover most scenarios for matting purpose.}
\label{fig:iteraction_eg}
\vspace{-10pt}
\end{figure}

\section{Architectures}
The network architecture of the virtual multi-modality foreground matting (VMFM) is designed to fully automatically extract accurate human-object interaction foreground instead of target selection through human-computer interaction. It first learns to estimate the same foreground alpha matte through the dual network (SFPnet and DFPnet) of the Foreground Prediction (FP) module. Since predictions of SFPnet and DFPnet are naturally biased to different regions, \eg segmentation map is sensitive to the human body, while depth map focuses more on the human-object interactive area. we introduce Complementary Learning (CL) to learn more dependable elements between dual FP networks. The overall architecture of VMFM training network is shown in detail in Figure~\ref{fig:architecture}.

\subsection{Foreground Prediction}
{\bf Modality preprocessing.} Given an input RGB image, we use a depth estimation backbone network (D-Net~\cite{alhashim2018high}) to estimate its depth map and S-Net (Mask-RCNN~\cite{he2017mask}) to automatically segment the mask of human. Moreover, we follow~\cite{wang2020learning} to employ Hourglass~\cite{newell2016stacked} as the backbone (I-Net) for predicting human-object paired heatmaps. As shown in Figure~\ref{fig:iteraction_eg}, interaction heatmaps can provide semantic prior of human and his or her interactive objects. For human without any associated object, the peak of interaction heatmap generalizes to the center point of him or her. 

Due to the difference between segmentation and depth features, we build a dual network for foreground prediction (FP) - segmentation-based foreground prediction network (SFPnet) and depth-based foreground prediction network (DFPnet). SFPnet receives the raw RGB image $I$, segmentation mask $S$, and interaction heatmap $H$ as inputs, while DFPnet replaces $S$ with the depth map $D$. In the following, $SFPnet(I, S, H)$ and $DFPnet(I, D, H)$ are referred to as $FP^{m}$, $m\in \left \{ 1, 2 \right \}$. Both of the dual networks are built in encoder-to-decoder fashion, and the architecture of each is well-modularized to make sure easy substitution to other networks that can serve the same purpose. We encourage other fellow researchers in the community to replace modules in VMFM to achieve better performance. The dual network of FP module is pre-trained on the subset of human subjects in the Adobe Matting Dataset~\cite{xu2017deep}. 
The pre-training is supervised by minimizing the $L_{a}^{m}$ loss,
\begin{equation}
    L_{a}^{m}=\left \| a^{m}-a^{*} \right \|_{1} + \left \| 	\nabla(a^{m})-	\nabla(a^{*}) \right \|_{1}
\label{E1}
\end{equation}
where $a^{m}$ is the alpha matte output of $FP^{m}$, $m\in \left \{ 1, 2 \right \}$, $a^{*}$ is the ground truth alpha matte, and the gradient term is beneficial to remove the over-blurred alpha matte~\cite{zhang2019late}.

We adopt a two-stage training process for foreground prediction in our method which are regular supervision and self-supervision respectively. In the first stage, we train the FP network under supervision (with labeled data). The proposed self-supervision is achieved through complementary learning on unlabeled data in the second stage. In addition, we introduce a discriminator based on LS-GAN~\cite{mao2017least} to distinguish between fake composites and real images to improve the foreground matting network. For the generator $\mathcal{G}$ update in the first stage, the adversarial loss term is:
\begin{equation}
\begin{aligned}
    L_{\mathcal{G}1}^{m} = \mathbb{E}_{X,\bar{B}\sim p_{X,\bar{B}}}[&(\mathcal{D}(a^{m}F^{*}+(1-a^{m})\bar{B})-1)^{2}\\
    &+\lambda_{a}L_{a}^{m}+ \lambda_{com}L_{com}^{m}]
\end{aligned}
\label{E2}
\end{equation}
\begin{equation}
    L_{com}^{m}=\left \| a^{m}F^{*}+(1-a^{m})B^{*}-I \right \|_{1}
\label{E3}
\end{equation}
where $a^{m}=\mathcal{G}(X)$, $X$ comprises RGB image $I$ and its corresponding virtual modalities, $L_{com}^{m}$ denotes the composite loss, which is the absolute difference between the input RGB image $I$ and the predicted RGB image generated from the ground truth foreground $F^{*}$, the true background $B^{*}$ and the predicted foreground alpha matte $a^{m}$. $ L_{com}^{m}$ can regularize the network to follow the compositional operation, which further reduces the error of foreground alpha prediction. $\bar{B}$
is a given background for
generating a composited image seen by the discriminator $\mathcal{D}$.

The objective for $\mathcal{D}$ is:
\begin{equation}
\begin{aligned}
    L_{\mathcal{D}}^{m} = \,&\mathbb{E}_{X,\bar{B}\sim p_{X,\bar{B}}}[(\mathcal{D}(a^{m}F^{*}+(1-a^{m})\bar{B}))^{2}]\\
    &+\mathbb{E}_{I \in p_{data}}[(\mathcal{D}(I)-1)^{2}]
\end{aligned}
\label{E4}
\end{equation}

For the generator's update in the second stage, we minimize: 
\begin{equation}
\begin{aligned}
    L_{\mathcal{G}2}^{m} = \mathbb{E}_{X,\bar{B}\sim p_{X,\bar{B}}}[&(D(a_{}^{m}F^{*}+(1-a_{}^{m})\bar{B})-1)^{2}\\
    &+\lambda_{cl}L_{cl}^{m} ]
\end{aligned}
\label{E5}
\end{equation}
where $L_{cl}^{m}$ denotes a complementary learning constraint for self-supervision between the dual network of FP module. For the discriminator in Stage 2, we also minimize Eq.~\ref{E4}. The details of the Complementary Learning (CL) module are described in Section~\ref{Complementary Learning}.

\begin{figure}[t]
\centering
    \includegraphics[width=0.93\linewidth]{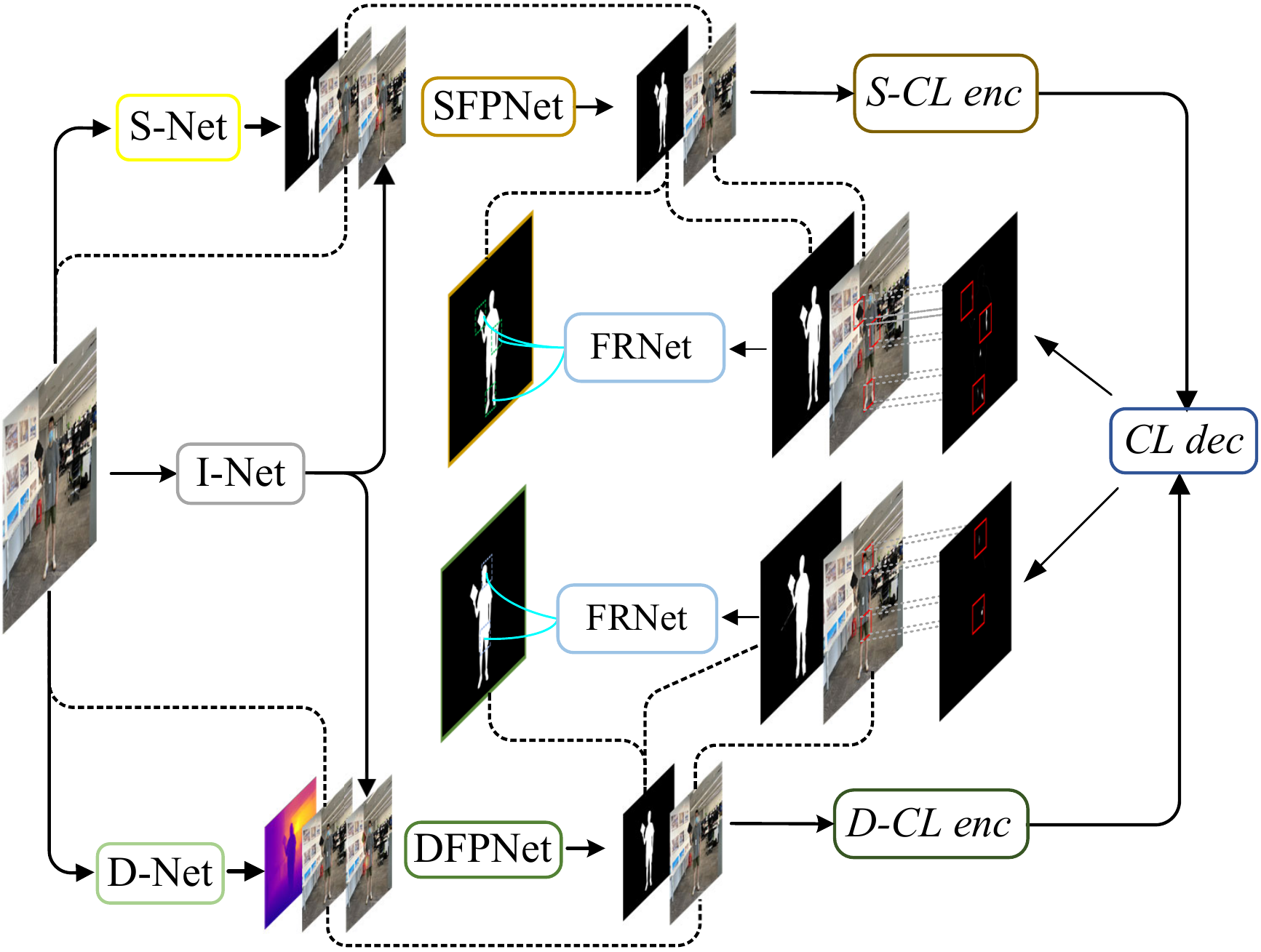}

\caption{Architecture of VMFM in the inference stage. {\bf FRNet:} foreground refinement network.}
\label{fig:inference}
\vspace{-6pt}
\end{figure}

\begin{figure*}[t]
\centering
    \includegraphics[width=0.95\linewidth]{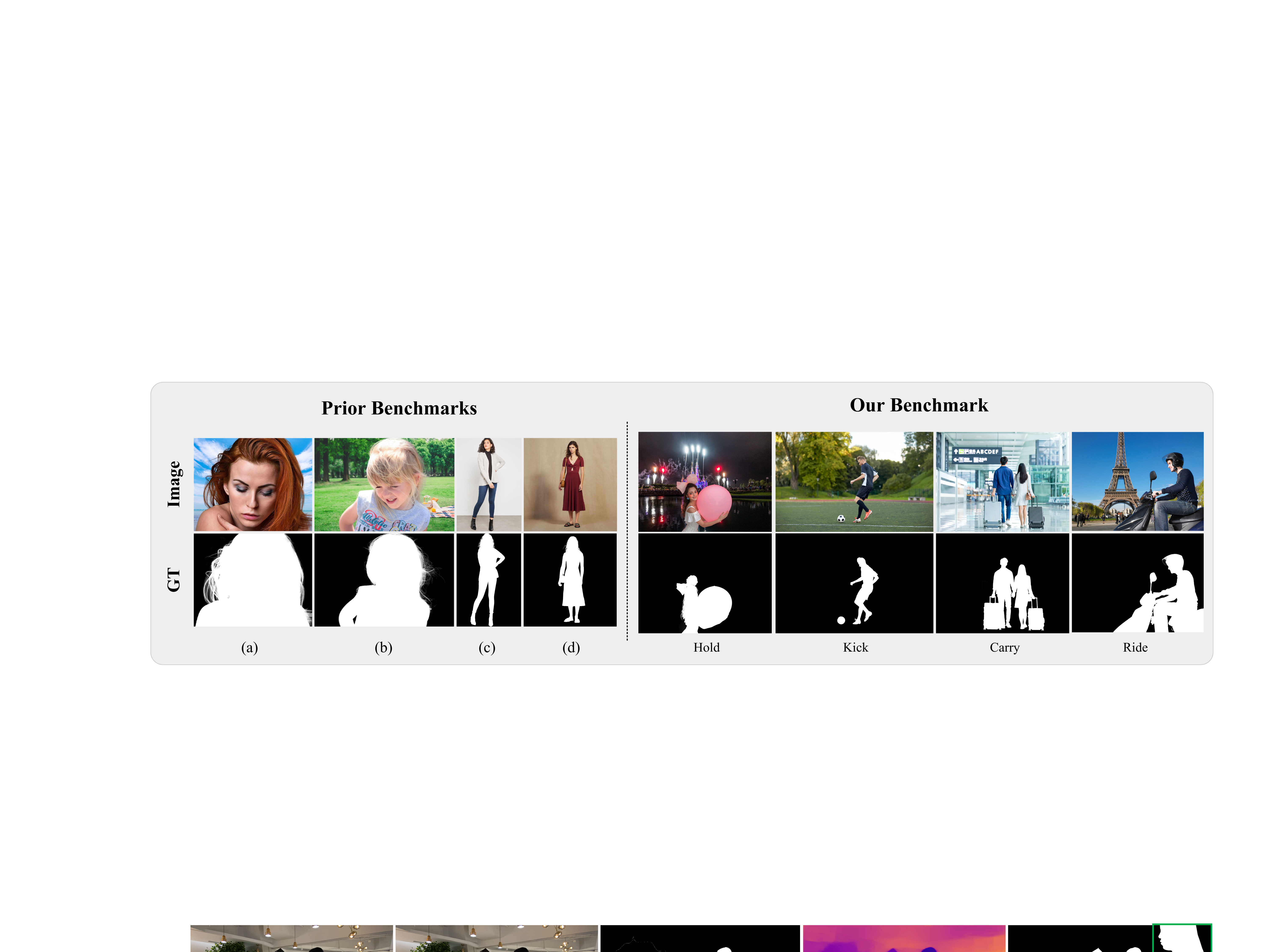}

\caption{Comparison of benchmarks. (a) and (b) are examples of the Adobe~\cite{xu2017deep} benchmark. (c) and (d) are examples of the Distinctions-646~\cite{Qiao_2020_CVPR} benchmark. Prior benchmarks mainly contain scenes of portrait and human-only, extremely rare human-object interaction. However, our benchmarks provide diverse classes of interactive objects, interactive modes and
scenarios to cover different matting application needs. 
}
\label{fig:comparison_benchmark}
\vspace{-8pt}
\end{figure*}
\subsection{Complementary Learning}\label{Complementary Learning}
We propose a dual network with different modalities to predict the same alpha matte. The underlying assumption is that each modality is good at predicting certain regions while weak in others. Therefore, we propose the Complementary Learning (CL) module to estimate the deviation probability map for predicted alpha mattes. The CL forces each of the dual network (SFPNet and DFPNet) to learn more reliable pixels between estimated matte $a^{1}$ and $a^{2}$ while rejecting the relatively unlikely one at the same location. We use encoder-to-decoder (enc2dec) mechanism which includes dual encoder $CL_{enc}^{m}(I, a^{m})$, $m\in \left \{ 1, 2 \right \}$ and one decoder $CL_{dec}(CL_{enc}^{m})$ in the CL module. 

{\bf CL module training.} We train the CL module simultaneously with the FP network training at the first stage, so that it is directly supervised by the true deviation probability map between each FP network result $a^{m}$ and its ground truth $a^{*}$. The loss is defined as
\begin{equation}
    \begin{split}
    L_{c}^{m}=\left \| CL(I^{m}, a^{m})-Q(|a^{m}-a^{*}|) \right \|_{1}
    \end{split}
\label{E6}
\end{equation}
where $CL(I^{m}, P^{m})$ denotes the CL module which is noted as $CL^{m}$ in the following, $Q$ denotes a dilation-normalization operation to broaden the area of complementary learning and produce a deviation probability map. We normalize the values in $CL^{m}$ to [0,1]. Interchangeably, we can also regard deviation probability as an expression of confidence. Intuitively one pixel with a higher probability value in the deviation probability map indicates lower confidence. We set a probability threshold $\tau \in (0,1)$ that only pixel with higher deviation probability than $\tau$ will be supervised by its counterpart (the pixel estimated by the other FP network at the same position) whose probability is lower than $\tau$. We update the higher value $CL_{i,j}^{m}$ (with lower confidence) at pixel $(i, j)$ to 1:   
\begin{equation}
    CL_{i,j}^{m} = \begin{Bmatrix}
1, \,if \ CL_{i,j}^{m} > \tau \\ 
CL_{i,j}^{m}, \,else
\end{Bmatrix}
\label{E7}
\end{equation}
{\bf Operation of complementary learning.} In the second stage (self-supervision) of foreground prediction (FP) training, we freeze the CL module and conduct complementary learning bounded by the constraint $L_{cl}^{m}$ in Eq.~\ref{E8} to update the dual FP network. 
\begin{equation}
    L_{cl}^{m} = \lambda_{cs}L_{cs}^{m}+\lambda_{dc}L_{dc}^{m}
\label{E8}
\end{equation}
where we introduce a complementary supervision constraint $L_{cs}^{m}$ to switch between SFPnet and DFPnet training:
\begin{equation}
\begin{split}
    L_{cs}^{m} &= \beta_{m}\left \| a^{m}-a^{3-m} \right \|_{1}\\
    \beta_{m}^{(i,j)} &= \begin{Bmatrix}
1, \,if \ CL_{i,j}^{m} > CL_{i,j}^{3-m} \\ 
0, \,else
\end{Bmatrix}
\end{split}
\label{E9}
\end{equation}
$\beta_{m}$ is a complementary learning area map, where $\beta_{m}^{(i,j)} = 1$ means that the current network $FP^{m}$ needs to learn from the result ($a_{i,j}^{3-m}$) of the other FP network ($FP^{3-m}$), at the pixel $(i,j)$ and vice versa, $m\in \left \{ 1, 2 \right \}$.

$L_{dc}^{m}$ is a deviation correction constraint to eliminate unreliably predicted pixels of $FP^{m}$:
\begin{equation}
\begin{split}
    L_{dc}^{m} &= \sigma_{m}\left \| CL^{m}-0 \right \|_{2}\\
     \sigma_{m}^{(i,j)} &= (CL_{i,j}^{1} \ is \ 1 \;\ and \;\ CL_{i,j}^{2} \ is \ 1)
\end{split}
\label{E10}
\end{equation}
$\sigma_{m}$ is a deviation correction area map, where $\sigma_{m}^{(i,j)} = 1$ means both pixels ($CL_{i,j}^{1}$ and $CL_{i,j}^{2}$) show low confidence. We set constraint $L_{dc}^{m}$ for updating dual FP network to limit the number of such unreliable pixel pairs. Obviously, $L_{cs}^{m}$ and $L_{dc}^{m}$ are non-interfering. 

\subsection{Foreground Refinement}
Although complementary learning improves the performance of foreground matting in the unlabeled condition, the encoder-to-decoder structure in FP module may over-regularize the results. Therefore, during the inference stage, we extend the above proposed pipeline by adding the matting refinement network ({\bf RN}), as illustrated in Figure~\ref{fig:inference}. To reduce computational complexity, we selectively extract top-$K$ pixels with the highest estimated errors in the deviation probability map as centers to define 16 $\times$ 16 patches in the predicted alpha matte for further refinement. Each selected patch is concatenated with its RGB region before being fed into the extended network. We apply loss $L^{m}_{a}$ (Eq.~\ref{E1}) to supervise the matting refinement network with the parameters of FP module fixed. We illustrate the effect of our refinement network in Section~\ref{experiment}. More network architecture details are given in Figure~\ref{FP_details} to~\ref{RN_details}.

\section{Experiments}\label{experiment}
We first describe the datasets used for training and testing. Subsequently, we compare our results with existing state-of-the-art (SOTA) foreground matting algorithms. Finally, We conduct ablation experiments to show the effectiveness of each module. More implementation details are provided in Appendix~\ref{Imp_Details}. 



\subsection{Datasets}
{\bf Pre-train.} We follow the method of~\cite{sengupta2020background} to produce 26.9k composited images for pre-training our dual Foreground Prediction (FP) network: 269 Adobe~\cite{xu2017deep} samples of human subjects composited over 100 random COCO~\cite{lin2014microsoft} images as backgrounds.
\begin{table}[t]
\renewcommand\tabcolsep{3.0pt}
\begin{center}
\scalebox{0.9}{
\begin{tabular}{ccccc}
\toprule
Dataset& Human-only & Interaction & $B^{*}$  & Annotation  \\
\midrule
DAPM~\cite{shen2016deep}&\checkmark&-&-&\checkmark\\
Adobe~\cite{xu2017deep}&\checkmark&25,3&-&\checkmark\\
SHM~\cite{chen2018semantic}&\checkmark&-&-&\checkmark\\
BSHMCA~\cite{liu2020boosting}&\checkmark&-&-&\checkmark\\
Dist-646~\cite{liu2020boosting}&\checkmark&-&-&\checkmark\\
BGM$_{real}$~\cite{sengupta2020background}&\checkmark&-&\checkmark&-\\
\midrule
Ours(LFM40K)&\checkmark&\checkmark&\checkmark&\checkmark\\
Ours(UFM75K)&\checkmark&\checkmark&\checkmark&-\\
\bottomrule
\end{tabular}}
\end{center}
\vspace{-10pt}
\caption{The configurations of foreground matting datasets. $B^{*}$: true background. (25,3) means that only about 25 foregrounds from the training set and 3 from the testing set in Adobe benchmark are interactive scenes by definition.}
\label{Datasets}
\vspace{-12pt}
\end{table}

Most of the existing matting datasets \eg~\cite{chen2018semantic,liu2020boosting,Qiao_2020_CVPR,xu2017deep} consist of HD SLR images, which focus narrowly on portrait or human body alone. However, these samples are still far different from mobile captures. More importantly, the diversity of human-object interaction is severely limited in these images. 

To facilitate the research in this area (especially the foreground matting in the human-object interactive settings), we construct two human-object interactive matting datasets which cover rich diversity of backgrounds. Our proposed model is trained accordingly to make sure good performance in real-world scenarios. The datasets and training strategy together significantly extend the use cases of the existing algorithms. We claim that our proposed model is bounded by none of the known constraints in the prior arts, such as trimap or predefined background. 

{\bf LFM40K.} We propose our first dataset Labeled Foreground Matting 40K ({\bf LFM40K}), which contains more than 40000 (31950 in the training set and 8050 in the test set) labeled frames (images) from 85 videos. LFM40K has 20 interaction classes, \eg carry, hold, kick, ski, work on computer and so on. We set 25\% of the dataset with human-only foreground scenes to better cover the rich diversity of different application scenarios.

{\bf UFM75K.} Our unlabeled dataset - Unlabeled Foreground Matting 75K ({\bf UFM75K}), consists of more than 75000 unlabeled images (62750 for self-supervised training and 12250 for testing) from 172 videos indoors and outdoors. Similarly, 25\% of the dataset is human-only foreground scene. UFM75K shares the same interactive classes with LFM40K but is more diverse in terms of interactive objects. We take full advantage of the inexpensive but large quantity of UFM75K under self-supervision to ensure the robustness of our network on different application scenarios. Interaction category and dataset details are introduced in Table~\ref{Interaction_classes}.  

Table~\ref{Datasets} and Figure~\ref{fig:comparison_benchmark} show the comparison between some existing foreground matting datasets with ours. Our datasets consist of both human-only and human-object interacted data. Our datasets could serve as a new challenging benchmark in the foreground matting area. LFM40K dataset is the first high-quality annotated human-object interactive dataset under unconstrained conditions while UFM75K covers a great diversity of real-world scenarios. 
\begin{table}[t]
\renewcommand\tabcolsep{4.5pt}
\begin{center}
\begin{tabular}{cccccc}
\toprule
{\bf Datasets}&\ {\bf Methods}&SAD& MSE& Grad& Conn\\
\midrule
\multirow{6}{*}{Adobe} &CAM&13.92& 4.67&70.32&10.02 \\

&IM& 14.61& 4.92&75.68&10.61\\
&LFM&40.35& 16.80 & 125.77 & 33.65\\
&BGM&14.12& 4.79&72.45&10.18 \\
\cdashline{2-6}[0.8pt/2pt]
&S-CL-RN$^{*}$&{\bf 12.68}&{\bf 4.39}&{\bf 62.91}&{\bf 8.64}\\
&D-CL-RN$^{*}$&13.01&4.68&71.23&9.02\\
\midrule
\multirow{6}{*}{Dist-646} & CAM& 17.40&5.11 &82.29&15.33\\

&IM&18.69& 5.82&88.61&17.26\\
&LFM&38.31& 15.63&112.75&36.12\\
&BGM& 12.82& 4.95&70.81&11.97\\
\cdashline{2-6}[0.8pt/2pt]
&S-CL-RN$^{*}$&{\bf 10.62}&{\bf 3.97}&{\bf 65.39}&{\bf 10.65}\\
&D-CL-RN$^{*}$&12.35&4.24&68.23&11.04\\
\bottomrule
\end{tabular}
\end{center}
\vspace{-10pt}
\caption{The quantitative results on composition benchmarks. We scale MSE, Grad and Conn by $10^{-2}$, $10^{2}$ and $10^{3}$.{\bf CAM:} Context-Aware Matting~\cite{hou2019context}, {\bf IM:} Index Matting~\cite{lu2019indices}, {\bf BGM:} Background Matting~\cite{sengupta2020background}, {\bf LFM:} a Late Fusion Matting~\cite{zhang2019late}. $^{*}$ means our proposed models. All competitive methods need additional inputs of trimaps (CAM and IM) or the true backgrounds (BGM), except LFM and ours.}
\label{Results_composite}
\vspace{-10pt}
\end{table}


\subsection{Comparative study on composition datasets}

We conduct comparative study on two composition benchmarks: Adobe~\cite{xu2017deep} and Dist-646~\cite{Qiao_2020_CVPR} datasets. To construct test benchmarks, we separately composite 11 held-out samples of human subjects from Adobe test set and 10 held-out ones of human subjects from Dist-646 with 20 random backgrounds per sample.  

\begin{figure*}[t]
\centering
    \includegraphics[width=1.0\linewidth]{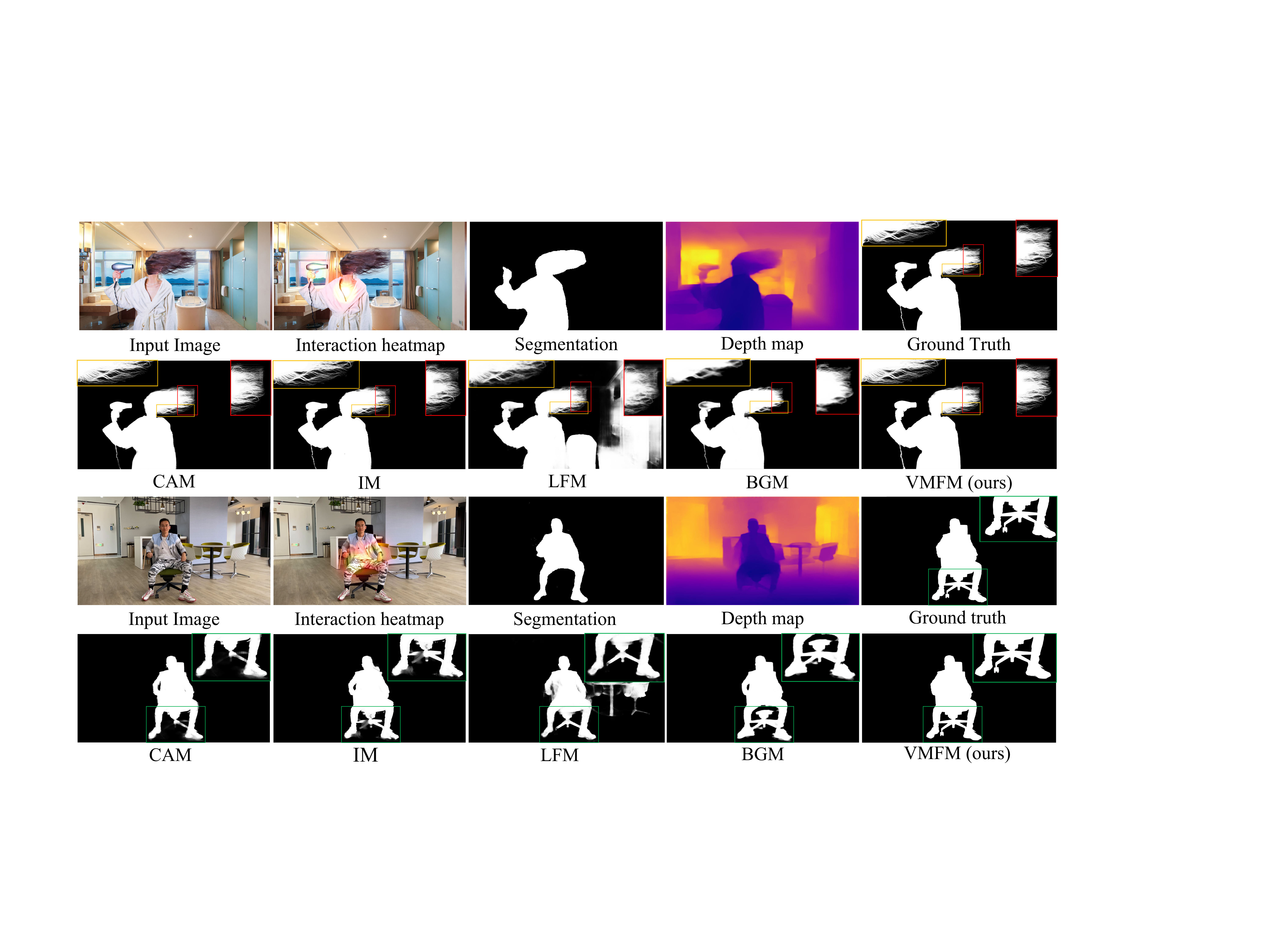}

\caption{Comparison of foreground matting methods.}
\label{fig:results}
\vspace{2.5pt}
\end{figure*}

We report mean square error (MSE), sum of the absolute difference (SAD), spatial-gradient (Grad) and connectivity (Conn) between predicted and ground truth alpha mattes. Lower values of these metrics indicate better estimated alpha matte. The quantitative results are shown in Table~\ref{Results_composite}. Although almost all test samples in both two test composition benchmarks are human-only scenes, our VMFM method (S-CL-RN) shows significant superiority over competing trimap-free method (LFM~\cite{zhang2019late}), and also outperforms trimap-based (CAM~\cite{hou2019context}, IM~\cite{lu2019indices}) and background-based (BGM~\cite{sengupta2020background}) ones. Moreover, VMFM can leverage more semantic prior from virtual modalities instead of additional inputs. Row 1 and 2 of Figure~\ref{fig:results} visualize comparisons on Adobe testing sample and more representative visualizations are provided in Figure~\ref{fig:visual_Adobe}. Effectiveness of each component in VMFM is detailed in Section~\ref{compare_lfm40k} and~\ref{ablation}. 
\begin{table}[t]
\renewcommand\tabcolsep{2.3pt}
\begin{center}
\begin{tabular}{ccccc}
\toprule
{\bf Methods}\qquad & SAD \qquad &MSE($10^{-2}$) \qquad&Grad($10^{2}$)\qquad&Conn($10^{3}$)\\
\midrule
CAM& 14.48& 5.01&81.62&18.34\\
IM&16.36& 6.73&96.49&20.96\\
LFM&43.29&15.83 &126.31&40.37\\
BGM& 10.06& 4.28&70.26&14.49\\
\midrule
SFPNet&18.28&8.05&105.91&21.23\\
S-CL$_{cs}$&11.85&4.29&78.63&15.70\\
S-CL$_{dc}$&16.43&6.81&102.35&21.65\\
S-CL&7.02&3.65&62.72&10.91\\
S-CL-RN&5.23&3.04&56.42&6.35\\
\cdashline{1-5}[0.8pt/2pt]
DFPNet&16.05&6.36&90.12&20.51\\
D-CL$_{cs}$&9.39&4.13&69.81&12.76\\
D-CL$_{dc}$&15.24&5.07&87.09&18.22\\
D-CL&5.20&2.94&56.07&6.29\\
D-CL-RN&{\bf 4.39}&{\bf 2.08}&{\bf 51.62}&{\bf 5.26}\\
\bottomrule
\end{tabular}
\end{center}
\vspace{-10pt}
\caption{The quantitative results on the test set of LFM40K. {\bf S-CL$_{cs}$ (D-CL$_{cs}$):} SFPNet (DFPNet) adds CL with only $L_{cs}$ constraint, {\bf S-CL$_{dc}$ (D-CL$_{dc}$):} SFPNet (DFPNet) adds CL with only $L_{dc}$ constraint.}
\label{Results_LFM40K_i}
\vspace{-8pt}
\end{table}
\subsection{Comparative study on LFM40K}\label{compare_lfm40k}
We re-scale all images to 512 $\times$ 512 pixel patches in training and testing of LFM40K dataset. Table~\ref{Results_LFM40K_i} compares our dual foreground matting (FP) network (SFPnet and DFPnet) with previous state-of-the-art (SOTA) foreground matting algorithms on diverse human-object interactive scenes. Because both {\bf IM} and {\bf CAM} require trimaps as additional inputs, we dilate-erode the ground truth alpha mattes for each image in LFM40K to form such trimaps.

The quantitative results on LFM40K are shown in Table~\ref{Results_LFM40K_i}. We observe that before applying CL, our SFPNet not surprisingly underperforms by small margin compared to CAM, IM, and BGM due to removal of trimap and background image, both of which require expensive manual creation and captures. Compared with LFM, both SFPNet and DFPNet exhibit significant superiority. That is because we introduce a virtual multi-modality mechanism to extend semantic space of the RGB image so that the proposed models can better capture the interactive information between human and objects. Our DFPNet is slightly inferior to CAM but superior to IM, which demonstrates that depth modality can better associate human body and the interacted object. After applying complementary learning, our methods (D-CL and S-CL) significantly outperform all previous SOTA foreground matting algorithms.

\subsection{Comparative study on UFM75K}
To evaluate and benchmark our algorithm on unlabeled data (UFM75K), we follow~\cite{sengupta2020background} to create the pseudo trimaps for trimap-based algorithms. The trimap creation implement details are described in Appendix~\ref{trimap_generation}. We compare our algorithm with other competing algorithms on UFM75K dataset for evaluation. Due to lack of ground truth alpha mattes, we composite the estimated mattes over a green or blue background and perform a psycho-physical user study that aligns with~\cite{sengupta2020background}. Since IM~\cite{lu2019indices} and our VMFM only estimate alpha matte (without generating foreground image $F$), we set $F = I$ for these algorithms.
\begin{table}[t]
\renewcommand\tabcolsep{3.0pt}
\begin{center}
\scalebox{0.87}{
\begin{tabular}{cccccc}
\toprule
D-CL-RN vs.& much better & better & similar & worse & much worse \\
\midrule
S-CL-RN&11.8\%&22.4\%&57.0\%&6.2\%&2.6\%\\
LFM&60.7\%&26.4\%&12.9\%&0\%&0\%\\
BGM&51.5\%&39.2\%&9.3\%&0\%&0\%\\
CAM&32.8\%&43.0\%&20.1\%&4.1\%&0\%\\
IM&35.6\%&46.5\%&14.0\%&3.9\%&0\%\\
\bottomrule
\end{tabular}}
\end{center}
\vspace{-10pt}
\caption{User Study: compare D-CL-RN with our S-CL-RN and other competing algorithms on UFM70K.}
\label{D_vs}
\vspace{-5pt}
\end{table}

{\bf User study.} We compare our composited videos head-to-head with each of the competing methods on 30 test videos from UFM75K. Similar to~\cite{sengupta2020background}, we present each user with a web page showing the raw video, our composite, and a competing composite; we randomly shuffle the order of the last two. Users were then asked to rate composite A relative to B on a scale of 1-5 (1 being ‘much worse’, 5 ‘much better’). Each video pair was rated by 20 users or more. 

The subjective metrics of user study shown in Table~\ref{D_vs} show clearly that our foreground matting model D-CL-RN outperforms our S-CL-RN and all other competing methods by significant margin. Benefiting from the unique strength of each modality and complementary improvement across modalities, our VMFM method can capture more accurate semantic information of real images. Some representative visualizations are provided in Row 3 to 4 in Figure~\ref{fig:results}.


\subsection{Ablation Experiments}\label{ablation}
{\bf Complementary Learning.} We evaluate complementary learning (CL) through dual constraint - $L_{cs}^{m}$ and $L_{dc}^{m}$. Since two constraints are independent, we prune Virtual Multi-modality Foreground Matting (VMFM) network with one single CL constraint ($L_{cs}^{m}$ or $L_{dc}^{m}$). Table~\ref{Results_LFM40K_i} shows the individual performance of $L_{cs}^{m}$, $L_{dc}^{m}$ and synergy of both when evaluating $FP^{m}$ network (SFPNet or DFPNet) on the LFM40K test set. The CL constraint $L_{cs}^{m}$ which aims to exchange confidence score between cross-modality outputs, significantly benefits the alpha matting performance. Also, $L_{dc}^{m}$ plays as an indispensable role in optimizing the collective performance of two modalities. Additionally, we demonstrate the performance gain after combining dual CL constraints.

{\bf Refinement.} The quantitative results of matting refinement are shown in Table~\ref{Results_LFM40K_i}. Improvement in all quantitative metrics demonstrates that the refinement network (RN) is the last critical process that further enhances and sharpens the estimated alpha mattes. Some visualizations are provided in Figure~\ref{fig:visual_ufm75k} to~\ref{fig:visual_ufm75k2}.  
\vspace{4.3pt}
\section{Conclusion}

In this paper, we present a matting technique that can extract high quality human-object interactive alpha mattes from a single RGB image. The training is self-supervised using our proposed complementary learning strategy which removes the bottleneck of gathering expensive alpha labels. Our method avoids using additional inputs, \eg a green screen background, extra captured real background, or a manual trimap. Extensive experiments demonstrate that our model outperforms current state-of-the-art algorithms, not with more but fewer inputs. For future works, it may be possible to extend the human-object interactive method to 3D body reconstruction, which still only focuses on body voxels.

\clearpage
  \appendix
  \renewcommand{\appendixname}{Appendix~\Alph{section}}

\section{Technical Details}
\subsection{Implementation Details}\label{Imp_Details}
\vspace{-5pt}
The training details and all hyper parameters are outlined in Table~\ref{implementation_details}. After pre-training on the Adobe dataset~\cite{xu2017deep}, we first train the dual FP network on labeled data (LFM40K) for 10 epochs (Sup$^{1}$). Then we continue the training of FP and activate the training of CL module which is supervised by the deviation probability map of each FP network, for another 30 epochs (Sup$^{2}$) . To present ablation study, we completed self-supervised training 3 times, each time under different setting: completed CL self-supervision ($L_{cs}$ and $L_{dc}$), one single supervision $L_{cs}$ and one single supervision $L_{dc}$, each for 25 epochs.

\begin{figure}[t]
\centering
    \includegraphics[width=0.9\linewidth]{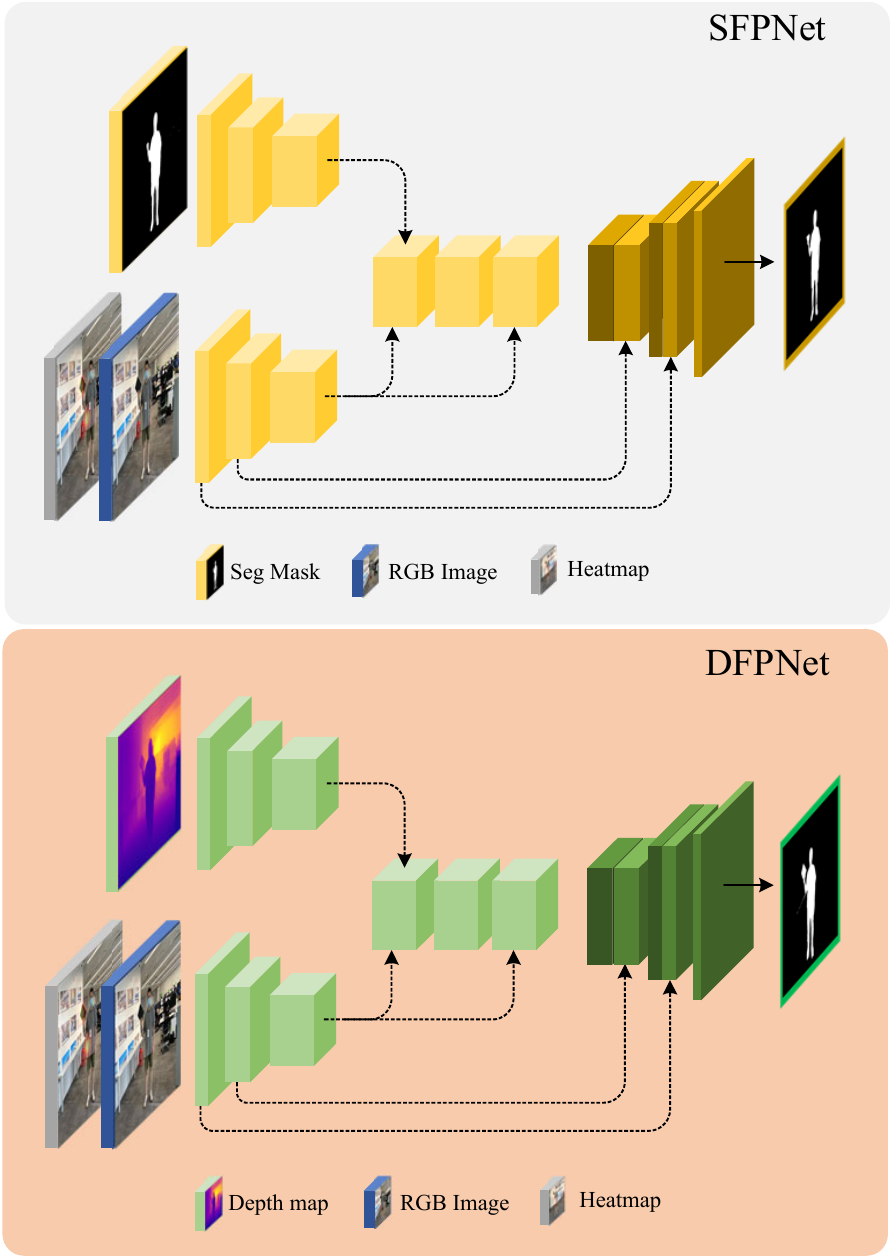}

\caption{Architecture of the dual FP network.} 
\label{fig:FPnet}
\vspace{-10pt}
\end{figure}

Architectures of SFPNet and DFPNet in dual foreground Prediction (FP) network are shown in Figure~\ref{fig:FPnet}. SFPnet receives raw RGB image $I$, interaction heatmap $H$ and segmentation mask $S$ as inputs, while DFPnet replaces $S$ with the depth map $D$. In each single FP network (SFPNet or DFPNet), we encoder input image (concatenated with the interaction heatmap) and the extra virtual modality ($S$ or $D$) respectively. Then the image feature vector and the extra virtual modality feature vector are fused within 3 convolution layers, followed by a decoder which outputs the estimated alpha matte. We select top-$k$ pixels with the highest estimated errors in the deviation probability map as centers to define 16 $\times$ 16 patches in the predicted alpha matte. Each selected patch is concatenated with its RGB region and then fed into the refinement network. We report the network architecture details in Table~\ref{FP_details}  to~\ref{RN_details}.

\begin{table}[t]
\renewcommand\tabcolsep{3.5pt}
\begin{center}
\scalebox{0.9}{
\begin{tabular}{lr}
\toprule
Parameter & Value  \\
\midrule
Optimizer \qquad\qquad\qquad\qquad\qquad\qquad\qquad & Adam \\
Learning rate & $1.0 \times 10^{-4}$\\
Number of Pre epochs & 20\\
Number of Sup$^{1}$ epochs & 10\\
Number of Sup$^{2}$ epochs & 30\\
\cdashline{1-2}[0.8pt/2pt]
Number of CL epochs & 25\\
Number of CL(w/o $L_{dc}$) epochs & 25\\
Number of CL(w/o $L_{cs}$) epochs & 25\\
\cdashline{1-2}[0.8pt/2pt]
Number of refinement epochs & 20\\
\cdashline{1-2}[0.8pt/2pt]
Batch Size & 4\\
Loss weight $\lambda_{a}$ & 1\\
Loss weight $\lambda_{com}$ & 0.5\\
Loss weight $\lambda_{cl}$ & 1\\
Loss weight $\lambda_{cs}$ & 6\\
Probability threshold $\tau$ & 0.5\\
Input image/extra modality size & $512 \times 512$\\
\bottomrule
\end{tabular}}
\end{center}
\vspace{-10pt}
\caption{Implementation details and hyper-parameter setting.}
\label{implementation_details}
\vspace{-18pt}
\end{table}

\subsection{Trimap Generation for Human-Object Interactive Scene}\label{trimap_generation}

To evaluate and benchmark our algorithm on unlabeled data (UFM75K), we need to form trimaps for trimap-based algorithms. We follow~\cite{sengupta2020background} to create the pseudo trimap and apply~\cite{chen2018encoder} to produce human segmentation $S_{h}$. Then we label each pixel with person-class probability $>$ 0.95 as foreground $F_{h}$, $<$ 0.05 as background $B_{h}$, and the rest as unknown area $U_{h}$. Similarly, we manually circle areas to cover the interacted objects and correspondingly label them as unknown area $U_{o}$. Finally, the human-object interactive trimap can be generated by Eq.~\ref{E0}:

\begin{equation}
\begin{aligned}
    F_{h\cup o} &= F_{h}\\
    U_{h\cup o} &= U_{h} \cup U_{o} - F_{h\cup o}\\
    B_{h\cup o} &= I_{area} - F_{h\cup o} \cup U_{h\cup o}
\end{aligned}
\label{E0}
\end{equation}
where $F_{h\cup o}$, $B_{h\cup o}$, $U_{h\cup o}$ denote foreground, background and unknown area of interaction trimap, $I_{area}$ denote the whole area of raw image $I$.

\subsection{Interaction Category}
The set of human-object interaction and corresponding object categories are listed in Table~\ref{Interaction_classes}. We follow the definition of interaction as in~\cite{gupta2015visual} and label 20 interaction classes to cover most human-object interaction application scenarios. In addition, human may have multiple interactions with a given object. LFM40K dataset is the first large-quantity and high-quality annotated human-object interactive matting dataset with diverse scenarios, which can facilitate other researchers in this area.  

\begin{table}[t]
\renewcommand\tabcolsep{3.5pt}
\begin{center}
\scalebox{0.95}{
\begin{tabular}{lr}
\toprule
Encoder (Image/vitual modality)\qquad\qquad & Output size  \\
\midrule
Conv+BN+ReLU & $64 \times 512 \times 512$\\
Conv+BN+ReLU & $128 \times 256 \times 256$\\
Conv+BN+ReLU & $256 \times 128 \times 128$\\
\midrule
\midrule
Fusion & Output size\\
\midrule
Conv+BN+ReLU & $256 \times 128 \times 128$\\
Conv+BN+ReLU & $256 \times 128 \times 128$\\
Conv+BN+ReLU & $256 \times 128 \times 128$\\
\midrule
\midrule
Deconv+BN+ReLU & $128 \times 256 \times 256$\\
Conv+BN+ReLU & $128 \times 256 \times 256$\\
Deconv+BN+ReLU & $64 \times 512 \times 512$\\
Conv+BN+ReLU & $64 \times 512 \times 512$\\
Conv+Tanh & $1 \times 128 \times 128$\\
\bottomrule
\end{tabular}}
\end{center}
\vspace{-10pt}
\caption{Network architecture for SFPNet and DFPnet.}
\label{FP_details}
\end{table}
\begin{table}[t]
\renewcommand\tabcolsep{3.5pt}
\begin{center}
\scalebox{0.95}{
\begin{tabular}{lr}
\toprule
Encoder ($CL^{m}_{enc}$)\qquad\qquad & Output size  \\
\midrule
Conv+BN+ReLU & $32 \times 256 \times 256$\\
Conv+BN+ReLU & $64 \times 128 \times 128$\\
Conv+BN+ReLU & $128 \times 64 \times 64$\\
Conv+BN+ReLU & $256 \times 64 \times 64$\\
\midrule
\midrule
Decoder ($CL_{dec}$) & Output size  \\
\midrule
Deconv+BN+ReLU & $128 \times 64 \times 64$\\
Conv+BN+ReLU & $128 \times 64 \times 64$\\
Deconv+BN+ReLU & $64 \times 128 \times 128$\\
Conv+BN+ReLU & $64 \times 128 \times 128$\\
Up-sample(2) & $64 \times 256 \times 256$\\
Conv+BN+ReLU & $32 \times 256 \times 256$\\
Up-sample(2) & $32 \times 512 \times 512$\\
Conv+BN+ReLU & $32 \times 512 \times 512$\\
Conv & $1 \times 512 \times 512$\\
\bottomrule
\end{tabular}}
\end{center}
\vspace{-10pt}
\caption{Network architecture for the complementary learning (CL) module.}
\label{CL_details}
\end{table}
\begin{table}[t]
\renewcommand\tabcolsep{3.5pt}
\begin{center}
\scalebox{0.95}{
\begin{tabular}{lr}
\toprule
Network\qquad\qquad\qquad\qquad & Output size  \\
\midrule
Conv+BN+ReLU & $k \times 4 \times 16 \times 16$\\
Conv+BN+ReLU & $k \times 8 \times 16 \times 16$\\
Conv+BN+ReLU & $k \times 8 \times 16 \times 16$\\
Conv+BN+ReLU & $k \times 4 \times 16 \times 16$\\
Conv+Tanh & $k \times 1 \times 16 \times 16$\\
\bottomrule
\end{tabular}}
\end{center}
\caption{Network architecture for the foreground refinement (RN) module.}
\label{RN_details}
\end{table}
\begin{figure}[h]
\centering
    \includegraphics[width=1.0\linewidth]{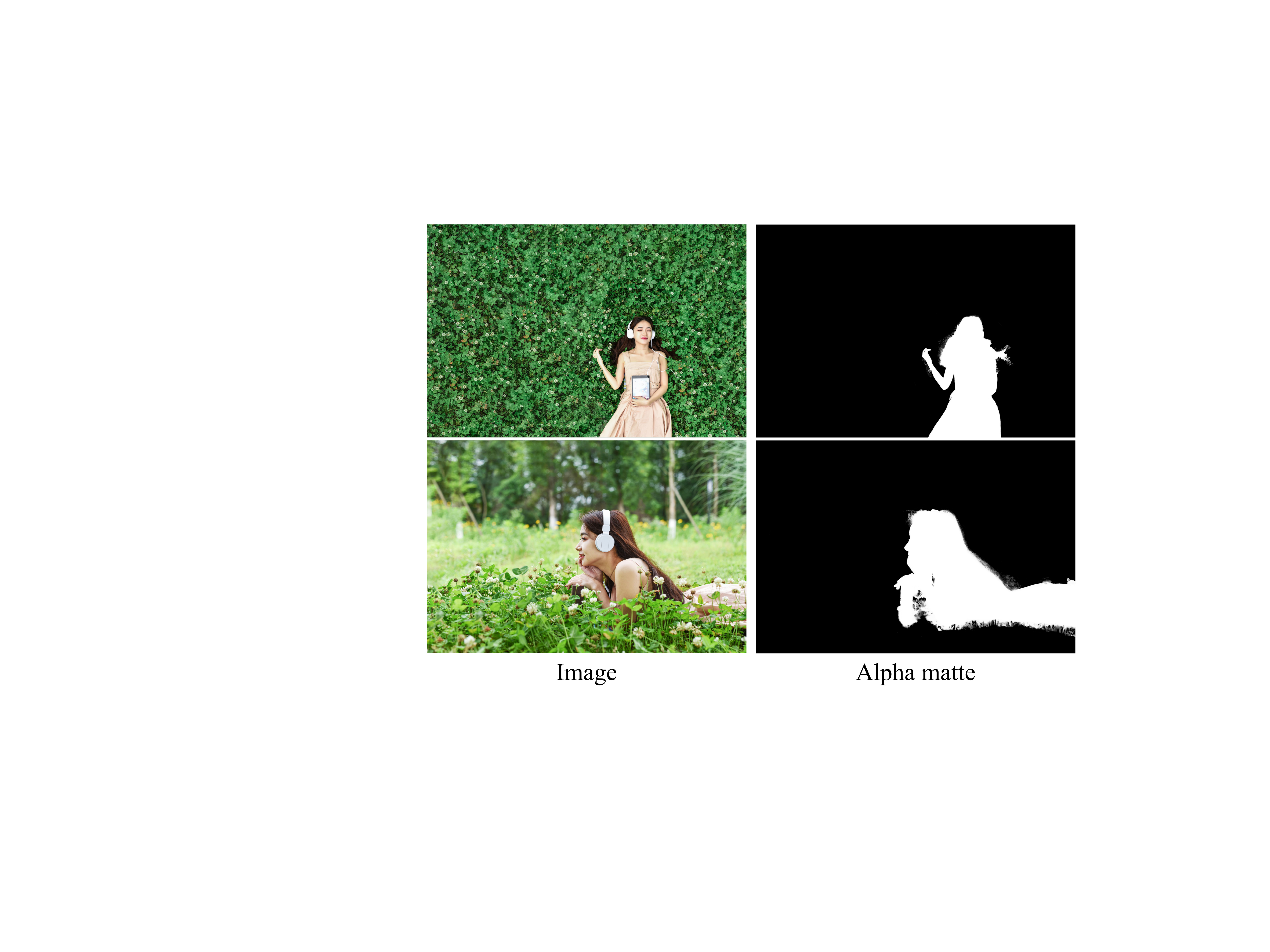}

\caption{Failure cases.} 
\label{fig:failure_case}
\vspace{-10pt}
\end{figure}
\section{More Visualization Results}

\begin{figure*}[t]
\centering
    \includegraphics[width=1.0\linewidth]{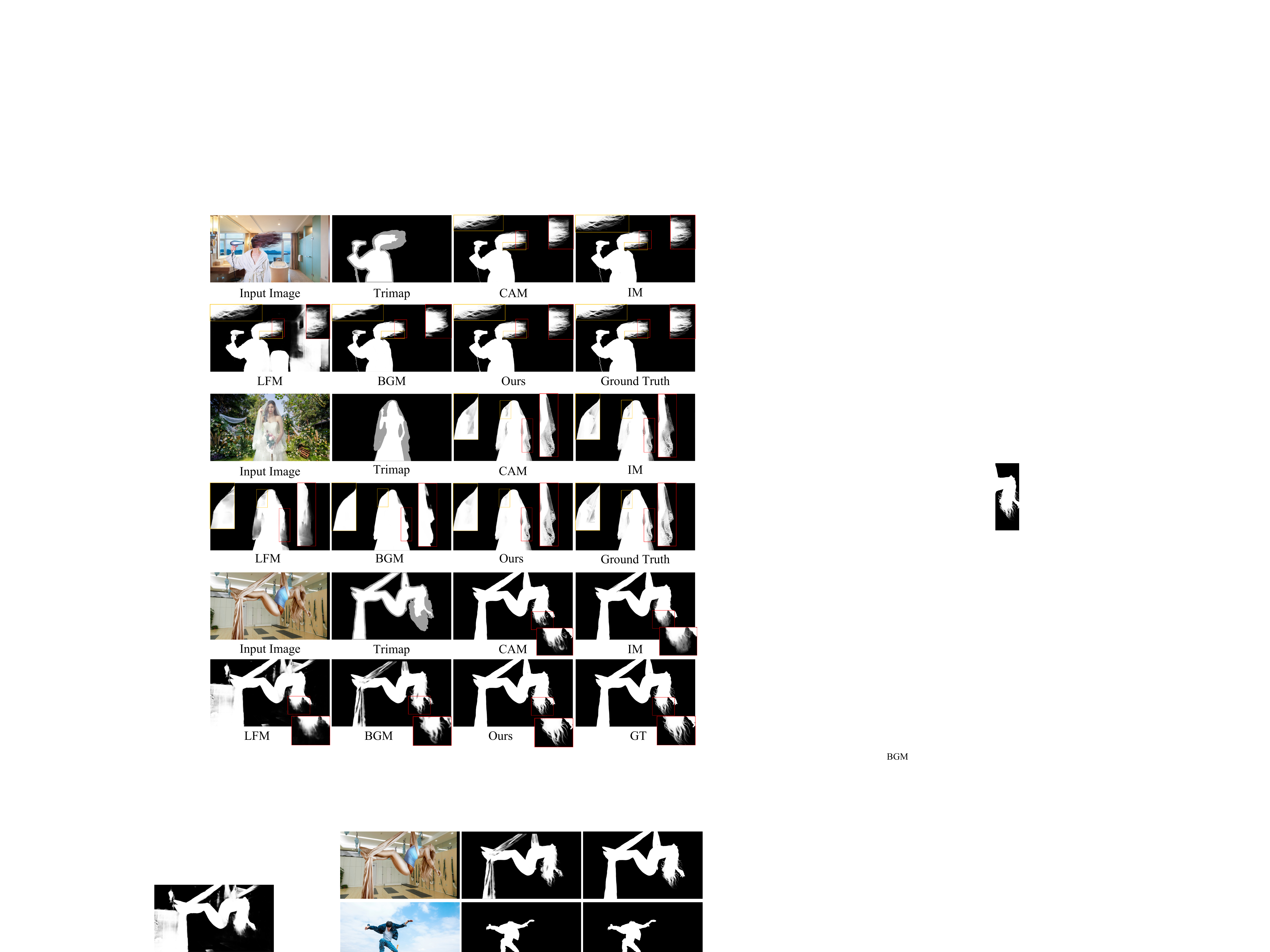}

\caption{Visual comparisons on the Adobe test benchmark.}
\label{fig:visual_Adobe}
\vspace{2.5pt}
\end{figure*}

{\bf Visual Results on Adobe dataset.} In Figure~\ref{fig:visual_Adobe} we show more visual comparisons on the Adobe test benchmark~\cite{xu2017deep}. Benefiting from the virtual multi-modality (depth and segmentation) and self-supervision (complementary learning), our method can capture semantic information of the human-object interactive images with higher degree of completeness. 

\begin{figure*}[t]
\centering
    \includegraphics[width=0.98\linewidth]{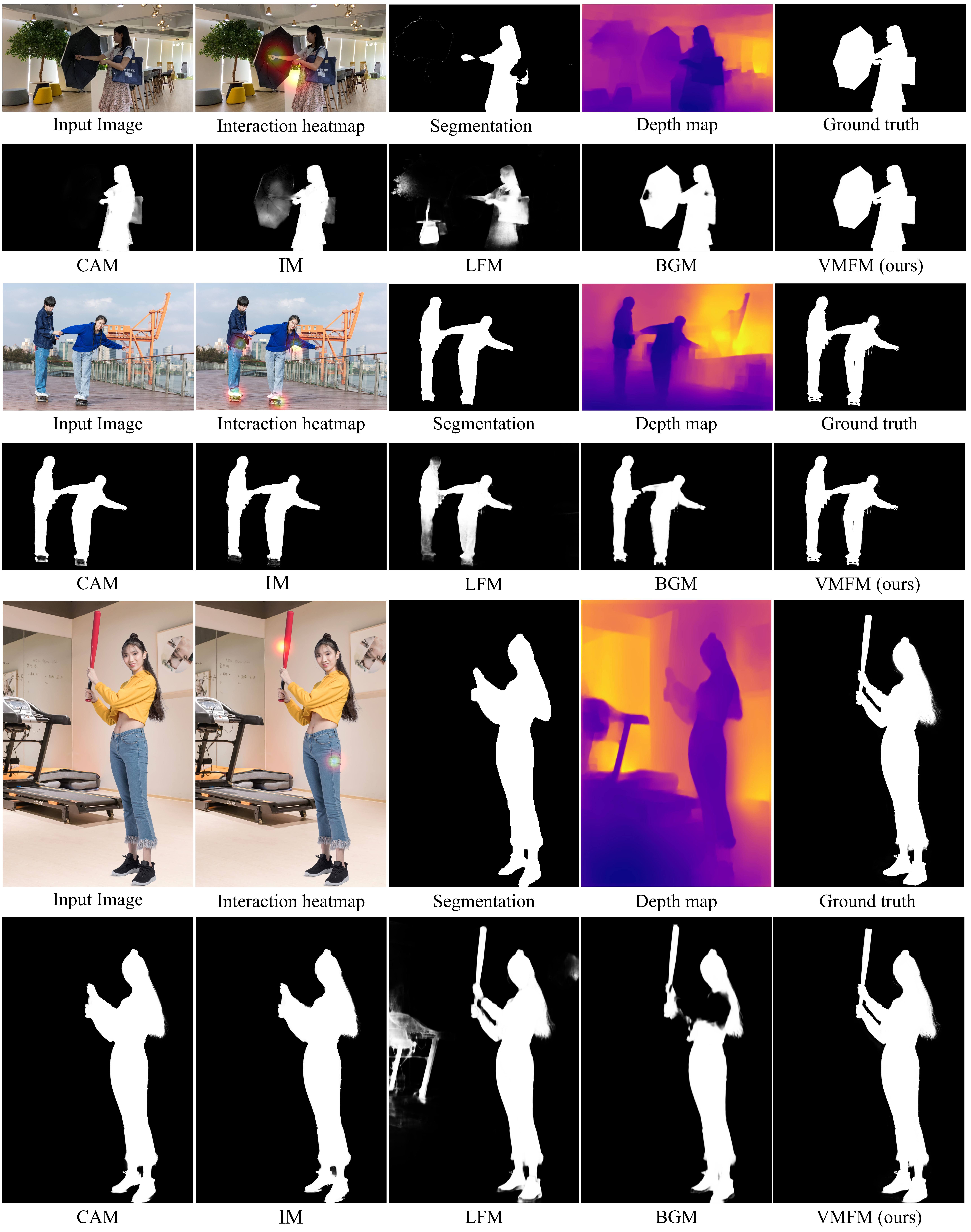}

\caption{Visual comparisons on UFM75K test set.}
\label{fig:visual_ufm75k}
\vspace{2.5pt}
\end{figure*}

\begin{figure*}[t]
\centering
    \includegraphics[width=0.96\linewidth]{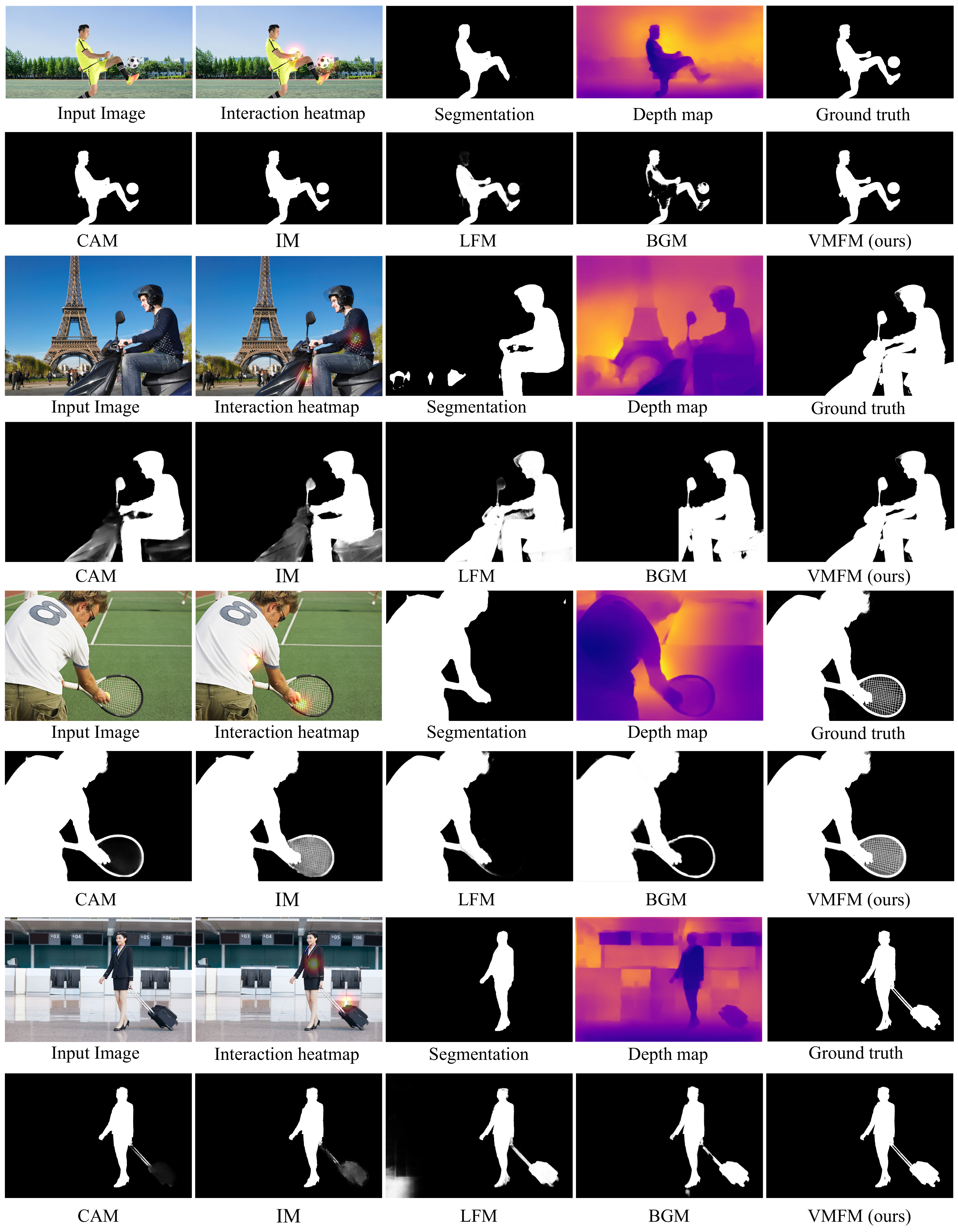}

\caption{Visual comparisons on UFM75K test set.}
\label{fig:visual_ufm75k2}
\vspace{2.5pt}
\end{figure*}

{\bf Visual Results on UFM75K dataset.} We also display more representative visualizations of our proposed VMFM on the UFM70K dataset. As illustrated in Figure~\ref{fig:visual_ufm75k} to~\ref{fig:visual_ufm75k2}, visual comparisons on real images further demonstrate the effectiveness and generalization of our algorithm in multiple scenarios. For future work, our foreground matting method may extend to cover a variety of real-world applications, \eg providing high-definition foreground mattes for 3D photo production, background replacement in live scene and film and TV show production, image editing and creation by personal computer or a mobile phone. We can also apply similar techniques to better interpret semantic information in the 3D body reconstruction under human-object interactive scene.

{\bf Limitation.} As shown in Figure~\ref{fig:failure_case}, our method is limited under the scenario that human and the background obscure each other. The algorithm performance is bounded in this case because the foreground and background are sophistically blended, which remains challenging for other state-of-the-art algorithms as well. However, we are working on solution to this kind of matting use case in future research. 

\clearpage

\begin{table*}[t]
\renewcommand\tabcolsep{3.5pt}
\begin{center}
\scalebox{0.95}{
\begin{tabular}{ccc}
\toprule
\multirow{2}{*}{Interaction class} &\multicolumn{2}{c}{Objects}\\
    \cmidrule(l){2-3} 
&\qquad\qquad\qquad\qquad LFM40K\qquad\qquad\qquad\qquad&\qquad\qquad\qquad\qquad UFM75K\qquad\qquad\qquad\qquad\\
\midrule
carry & handbag, backpack, umbrella, box & luggage, sports ball, hairdryer, box\\
&&handbag, backpack, umbrella\\
catch & sports ball, frisbee& sports ball, frisbee\\
cut & scissors, knife & fork, scissors, knife\\
drink &bottle& wine glass, cup, bowl\\
eat & hot dog, sandwich, banana & apple, orange, hot dog, cake\\
&&carrot, pizza, donut\\
hold & pen, cup, pad, box, flower& box, ball, phone, hairdryer\\
&&computer, bottle, flute, fan\\
hit & tennis racket, baseball bat & tennis racket, baseball bat, sports ball\\
jump & snowboard, skateboard, skis, surfboard & snowboard, skateboard, skis, surfboard\\
kick & sports ball & sports ball\\
lay & bench, dining table & bench, hammock, bed, couch,chair\\
read & book, pad & book, pad\\
ride & motorcycle & bicycle, motorcycle\\
sit & bench, chair & couch, bed, dining table, suitcase\\
skateboard & skateboard & skateboard\\
ski & ski & ski\\
snowboard & snowboard & snowboard\\
surf & surfboard & surfboard\\
talk on phone & cell phone & cell phone\\
throw & sports ball, frisbee & sports ball, frisbee\\
work on computer & laptop & laptop\\ 


\bottomrule
\end{tabular}}
\end{center}
\vspace{-10pt}
\caption{Interaction category and corresponding objects.}
\label{Interaction_classes}
\vspace{-12pt}
\end{table*}
\end{document}